\documentclass[conference]{IEEEtran}
\IEEEoverridecommandlockouts
\usepackage{cite}
\usepackage{amsmath,amssymb,amsfonts}
\usepackage{algorithmic}
\usepackage{graphicx}
\usepackage{textcomp}
\usepackage{balance}
\usepackage{xcolor}
\def\BibTeX{{\rm B\kern-.05em{\sc i\kern-.025em b}\kern-.08em
    T\kern-.1667em\lower.7ex\hbox{E}\kern-.125emX}}
\begin{document}

\title{Graph Contrastive Learning via Cluster-refined Negative Sampling for Semi-supervised Text Classification\\
}

\author{
	\IEEEauthorblockN{Wei Ai}
	\IEEEauthorblockA{\textit{College of Computer and Mathematics} \\
		\textit{Central South University of Forestry and Technology}\\
		ChangSha, China \\
		weiai@csuft.edu.cn}
	\and
	\IEEEauthorblockN{Jianbin Li}
	\IEEEauthorblockA{\textit{College of Computer and Mathematics} \\
		\textit{Central South University of Forestry and Technology}\\
		ChangSha, China \\
		jianbinli@csuft.edu.cn}
	\and
	\IEEEauthorblockN{\hspace{5em}Ze Wang}
		\IEEEauthorblockA{\textit{\hspace{4em}College of Computer and Mathematics}\\
		\textit{\hspace{4em}Central South University of Forestry and Technology}\\
		\hspace{4em}ChangSha, China \\
		\hspace{4em}zewang@csuft.edu.cn}
	\and
	\IEEEauthorblockN{Jiayi Du}
	\IEEEauthorblockA{\textit{College of Computer and Mathematics} \\
		\textit{Central South University of Forestry and Technology}\\
		ChangSha, China \\
		dujiayi@csuft.edu.cn}
	\and
    \IEEEauthorblockN{\hspace{5em}Tao Meng{*}}
	\IEEEauthorblockA{\textit{\hspace{4em}College of Computer and Mathematics} \\
		\textit{\hspace{4em}Central South University of Forestry and Technology}\\
		\hspace{4em}ChangSha, China \\
		\hspace{4em}mengtao@hun.edu.cn}
	\and
	 \IEEEauthorblockN{\hspace{5em}Yuntao Shou}
	\IEEEauthorblockA{\textit{\hspace{4em}College of Computer and Mathematics} \\
		\textit{\hspace{4em}Central South University of Forestry and Technology}\\
		\hspace{4em}ChangSha, China \\
		\hspace{4em}shouyuntao@stu.xjtu.edu.cn}
	\and
	\IEEEauthorblockN{\hspace{8em}Keqin Li}
	\IEEEauthorblockA{\hspace{8em}Department of Computer Science \\
		\hspace{6em}State University of New York\\
		\hspace{6em}New Paltz, New York 12561, USA \\
		\hspace{9em}lik@newpaltz.edu}
	\thanks{* is the corresponding author.}
}

\maketitle

\begin{abstract}
Graph contrastive learning (GCL) has been widely applied to text classification tasks due to its ability to generate self-supervised signals from unlabeled data, thus facilitating model training. However, existing GCL-based text classification methods often suffer from negative sampling bias, where similar nodes are incorrectly paired as negative pairs. This can lead to over-clustering, where instances of the same class are divided into different clusters. To address the over-clustering issue, we propose an innovative GCL-based method of graph contrastive learning via cluster-refined negative sampling for semi-supervised text classification, namely ClusterText. Firstly, we combine the pre-trained model Bert with graph neural networks to learn text representations. Secondly, we introduce a clustering refinement strategy, which clusters the learned text representations to obtain pseudo labels. For each text node, its negative sample set is drawn from different clusters. Additionally, we propose a self-correction mechanism to mitigate the loss of true negative samples caused by clustering inconsistency. By calculating the Euclidean distance between each text node and other nodes within the same cluster, distant nodes are still selected as negative samples. Our proposed ClusterText demonstrates good scalable computing, as it can effectively extract important information from from a large amount of data. Experimental results demonstrate the superiority of ClusterText in text classification tasks.
\end{abstract}


\section{Introduction}
Text classification stands as a fundamental and essential task in natural language processing, drawing significant interest from both academic researchers and industry professionals \cite{ai2023multi, ai2024edge, shou2022conversational, shou2025masked, shou2023comprehensive, meng2023deep, meng2024multi, meng2024deep, shou2023low}. This field of study finds diverse applications, including topic categorization and sentiment analysis. The ability to learn text feature representations is the primary reason behind the high performance of text classification models. Deep learning methods, such as Convolutional Neural Networks \cite{wang2024graph, shou2024adversarial, ai2024gcn, ai2023gcn, ai2023two, shou2023graph, shou2023czl} and Recurrent Neural Networks\cite{wang2018sentiment}\cite{dieng2016topicrnn}, have been widely applied in the field of text classification. With the advent of the Transformer architecture, approaches based on pre-trained models have demonstrated exceptional performance in text classification tasks. Recently, there has been a rising trend in utilizing graph neural networks (GNNs) for text classification applications, owing to their capability to model complex  semantic and structural information by representing textual data as graph structures \cite{meng2024masked}\cite{ai2024gcn}\cite{chen2019gated}\cite{chen2020citywide}.

\begin{figure}
	\centering
	\includegraphics[width=0.4 \textwidth]{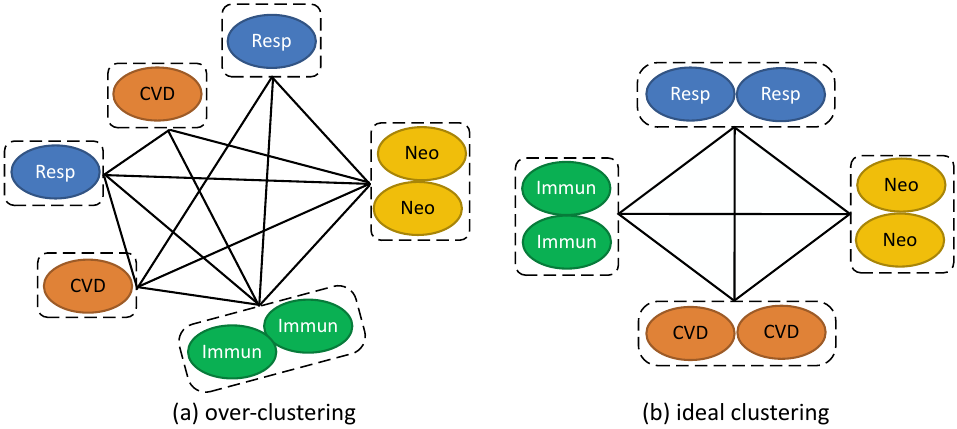}
	\caption{Illustration of over-clustering and ideal clustering. Different colors denote distinct text node categories.In (a), the CVD and Resp nodes are assigned to different clusters in the embedding space.}
	\label{tab:over-ideal}
\end{figure}

Graph-based text classification methods \cite{yao2019graph, shou2024contrastive, shou2024revisiting, meng2024masked, shou2024efficient, meng2024revisiting, shou2023graphunet}\cite{zhang2020text}\cite{lin2021bertgcn} have demonstrated remarkable performance by constructing the overall structure of the corpus. The advantage of these approaches lies in their ability to consider not only the content of individual documents but also the information propagated from adjacent nodes, leading to a more comprehensive determination of document categories. Traditional graph-based text classification methods take into account both labelled (training set) and unlabeled (test set) nodes during the information propagation process. However, when computing the cross-entropy loss, these approaches depend on labeled nodes, neglecting to exploit the abundant information embedded within numerous unlabeled nodes. Recent researches have incorporated graph contrastive learning into graph-based text classification to investigate the impact of unlabeled data in model training.

GCL can effectively utilize unlabelled node representations to compute contrastive loss, thereby optimizing the model without relying on true labels \cite{shou2024adversarial, shou2022object, ying2021prediction, shou2024contrastive, shou2024spegcl}. This approach significantly expands the range of data available for training and improves the model's generalization ability. However, despite the significant progress made by GCL-based text classification methods\cite{yang2022contrastive}\cite{zhao2023textgcl}\cite{lan2023contrastive}\cite{sun2022contrastive}, a critical issue remains in the negative sample selection process: existing methods may mistakenly select nodes with the same label as negative sample pairs. For example, CGA2TC\cite{yang2022contrastive} and TextGCL\cite{zhao2023textgcl} employ variations of InfoNCE as their contrastive loss function, considering identical nodes across views as positives while treating other nodes both within and across views as negative examples. This inevitably results in text nodes with shared labels being grouped as negative pairs, futilely increasing the dissimilarity between intra-class text nodes. As illustrated in Fig \ref{tab:over-ideal}, negative sampling bias forces text nodes belonging to the same class to be more distant in the embedding space, causing instances of the same class to be assigned to different clusters (i.e., over-clustering). Therefore, how to select the correct set of negative samples for anchor nodes to avoid over-clustering of text nodes remains an unresolved problem.

To address the issues mentioned above, we propose an innovative GCL-based method of graph contrastive learning via cluster-refined negative sampling for semi-supervised text classification. Specifically, we construct word-word and word-document relationships in the given corpus to form a heterogeneous text graph that retains global information. By dropping edges within the original text graph, we generate diverse contrastive views. Subsequently, we integrate the pre-trained model Bert and graph convolution networks to learn robust text representations. Next, we propose a clustering refinement strategy, which utilizes pseudo-labels generated by text clustering to optimize the negative sampling process in contrastive learning, selecting nodes from different clusters as the negative sample set for anchor nodes. Additionally, we present a self-correction mechanism to expand the number of correct negative samples within the same cluster. This helps to mitigate the negative impact caused by significant differences between clustering pseudo labels and true labels. ClusterText effectively alleviates the issue of over-clustering by optimizing the negative sample selection process. By integrating the scalable computational capabilities of GCL, ClusterText can effectively harness and utilize the rich data information. The contributions of this paper can be summarized as follows:

\begin{itemize}
	\item We propose a clustering refinement strategy to optimize the negative sampling process in graph contrastive learning.
	
	\item We present a self-correction mechanism, which selects nodes within the same cluster but distant from the anchor node as negatives, further refining the negative sample candidate set.
	
	\item We conduct experiments on five popularn benchmark datasets, and the results demonstrate the effectiveness of our proposed method for text classification tasks.
\end{itemize}

The remainder of this paper is organized as follows: Section \ref{sec:related_work} reviews related work, Section \ref{sec:proposed_method} provides a detailed description of the proposed method, Section \ref{sec:experrments} presents the experimental setup and results, and finally, Section \ref{sec:conclusion} offers a brief conclusion.

\begin{figure*}
	\centering
	\includegraphics[width=0.9 \textwidth]{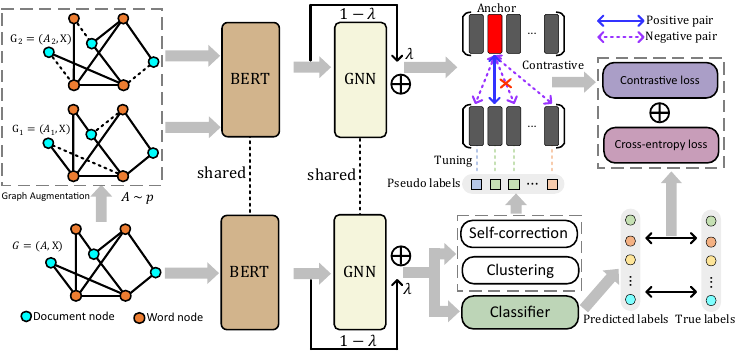}
	\caption{The overall process of the proposed ClusterText. The views are generated by applying augmentations to the original graph. Next, the representations learned from the original graph are clustered, and the clustering results are refined through self-correction to obtain pseudo-labels. These pseudo-labels are then used to optimize the negative sample selection process. Finally, the model is trained by combining contrastive loss and cross-entropy loss.}
	\label{fig:overall}
\end{figure*}

\section{RELATED WORK}
\label{sec:related_work}
\subsection{Graph-based Text Classification}
Due to the tremendous success of representation learning on graph-structured data, numerous innovative studies have emerged in recent years, making significant progress in improving the accuracy and efficiency of text classification. TextGCN\cite{yao2019graph} constructs a large heterogeneous text graph representing the entire corpus, including documents and words as nodes. By leveraging Graph Convolutional Networks (GCNs), TextGCN effectively extracts and learns the rich semantic relationships between texts. TextING\cite{zhang2020every} adopts a method of independently constructing a graph for each document, significantly reducing memory usage. This model uses Gated Graph Neural Networks (GGNN) to capture essential information in the text, thereby enhancing text classification performance. The BertGCN\cite{lin2021bertgcn} model combines the strengths of Graph Neural Networks and BERT pre-trained models, achieving significant performance improvements in transductive text classification. Overall, graph-based methods have demonstrated tremendous potential in text classification tasks.

\subsection{Graph Contrastive Learning}
Contrastive learning in the field of graph representation learning is known as Graph Contrastive Learning. MVGRL\cite{hassani2020contrastive} uses graph diffusion augmentation to simultaneously learn node and graph representations by contrasting nodes and graphs. Additionally, NCLA \cite{shen2023neighbor} proposes an innovative contrastive loss function that uses topological structure as a supervision signal to improve negative sampling bias. S3CL\cite{ding2023eliciting} enhances both inter-class distinction and intra-class similarity, resulting in the generation of high-quality node representations. For GCL-based text classification methods, CGA2TC\cite{yang2022contrastive} proposes an adaptive augmentation strategy that effectively reduces noise in text graphs and learns discriminative text representations. TextGCL\cite{zhao2023textgcl} learns richer text features by jointly training GNNs and large-scale pre-trained models. However, their negative sampling processes suffer from bias, and the erroneous negative pairs formed may lead to over-clustering issues.

\section{PROPOSED METHOD}
\label{sec:proposed_method}
In this section, we introduce the proposed ClusterText. The overall process of ClusterText is illustrated in Fig \ref{fig:overall}.

\subsection{Text Graph Construction}
The advantage of graph-based text classification methods lies in their ability to determine the text category by leveraging both the information within the document and the information from its neighboring nodes. We adhere to the established graph construction rules\cite{yao2019graph}, representing the corpus as a heterogeneous network encompassing both word and document nodes. Formally, the process for calculating edge weights between different nodes is as follows:
\begin{equation}
	A_{ij} = \begin{cases}
		\text{PMI}(i,j) & \text{if } i, j \text{ are words, PMI}(i,j) > 0 \\
		\text{TF-IDF}(i,j) & \text{if } i \text{ is document, } j \text{ is word} \\
		1 & \text{if } i = j \\
		0 & \text{otherwise}
	\end{cases}
\end{equation}

To fully utilize the attribute information, we use the pre-trained model Bert to initialize the representations of document nodes. Following the previous work BertGCN\cite{lin2021bertgcn}, we set the initial representations of word nodes to zero. The initial node feature matrix of the text graph can be represented as follows:
\begin{equation}
	X = \left( \frac{X_{\text{doc}}}{0} \right)_{(n_{\text{doc}} \times n_{\text{word}}) \times d}
\end{equation}
where $d$ is the dimension of the node representations.

\subsection{Text Graph Augmentation}
In graph contrastive learning, graph augmentation aims to improve the model's generalization ability and robustness by generating multiple transformed views of the graph, thereby helping the model learn more generalizable node representations. Current popular graph augmentation paradigms\cite{you2020graph}\cite{zhu2021graph}\cite{mo2022simple} typically perturb the original graph based on graph structure and node attribute perspectives. In our work, we employ edge dropping as the augmentation strategy by modifying the graph topology. Given the input adjacency matrix $A$ and the edge dropping probabilities, we perform graph augmentation to provide sufficiently diverse contrastive views for subsequent contrastive learning. In summary, our text graph augmentation process is as follows:
\begin{equation}
	G_1, G_2 \leftarrow \textbf{Aug}(A, p)
\end{equation}
where $G_1$, $G_2$ represent different augmented contrastive views, and $p$ is the edge dropping probabilities.

\subsection{Text Representation Learning}
In this paper, we adopt a standard two-layer GCN encoder. Given the input feature matrix \(X\) and the adjacency matrix \(A\), the computation process for node representations can be formalized as follows:
\begin{equation}
	H = \tilde{A}\sigma(\tilde{A}XW^{(0)})W^{(1)}
\end{equation}
where $\tilde{A} = \hat{D}^{-\frac{1}{2}} \hat{A} \hat{D}^{-\frac{1}{2}}$, $\hat{A} = A + I$. $I$ is the identity matrix. $W_0$ and $W_1$ correspond to the learnable parameters of the two-layer GCN. $\sigma$ is the nonlinear activation function.

Drawing inspiration from BertGCN\cite{lin2021bertgcn}, we integrate the learned text representations from Bert and GCN. The integration of text representations can be formulated as follows:
\begin{equation}
	Z = \lambda \textbf{FC}(H_{doc}) + (1 - \lambda)\textbf{FC}(X_{doc})
\end{equation}
where $H_{doc}$ represents the document node representations learned from the GCN. $\textbf{FC}$ is the fully-connected layers.

\subsection{Cluster-based negative sampling}
Current GCL-based text classification methods typically employ common negative sampling strategies. In particular, for any given text node $v_i$, the embedding learned in one augmented view is designated as the anchor, while the embeddings of $v_i$ across view acts as the positives. Concurrently, embeddings of other text nodes within these two augmented views are automatically chosen as negative samples. This strategy fails to consider the label information of the texts, leading to the occurrence of false negative pairs. According to the theory of GCL, false negative pairs can increase the distance between text nodes belonging to the same category, erroneously classifying them into different clusters and thus causing over-clustering.

To address the negative sampling bias mentioned above, we propose a clustering refinement strategy. This strategy employs unsupervised machine clustering methods to cluster text node embeddings learned from the original text graph. The text clustering process can be formalized as follows:
\begin{equation}
	C = \textbf{Cluster}(Z)
\end{equation}
where $Z$ denotes the fused text representations, and $C$ represents the pseudo labels of the texts. Based on the obtained pseudo labels $C$, we select the remaining nodes from different clusters as the negative sample set for each anchor node. Those nodes within the same cluster, which tend to share the true label with the anchor, are excluded from being selected as negative samples.

In the field of machine learning, achieving absolute precision in clustering algorithms remains a highly challenging task. If the pseudo labels significantly differ from the true labels, the desired negative candidate nodes within the same cluster (i.e., those that share the different true label as the anchor node) might be overlooked. This could prevent the model from learning the characteristic differences between inter-class samples. Therefore, we present a self-correction mechanism to expand negative samples within the same cluster. Specifically, we calculate the Euclidean distance within the nodes of the same cluster and set a hyperparameter $d$ to select the farthest $d\%$ from the anchor node as negative samples. It can be formulated as follows:
\begin{equation}
	dist_{ij} = \text{EuclideanDist}(v_i, v_j)
\end{equation}
\begin{equation}
	S_i = \{{dist_{ij} \mid v_j \in C(v_i), j \neq i}\}
\end{equation}
\begin{equation}
	dist_{ij}^{thr} = S_i^{sorted} [\lceil d\% \times n \rceil - 1]
\end{equation}
where $n$ is the number of nodes within the cluster. $S_i^{sorted}$ is an ascending order set. All nodes whose distances to the anchor exceed $dist_{ij}^{thr}$ are defined as distant nodes.
\begin{equation}
	distribution(v_i,d) = \{v_j \mid dist_{ij} > dist_{ij}^{thr} \}
\end{equation}
where $distribution(v_i,d)$ represents the distribution of the farthest $d\%$ nodes from the anchor node.

\begin{figure}
	\centering
	\includegraphics[width=0.45 \textwidth]{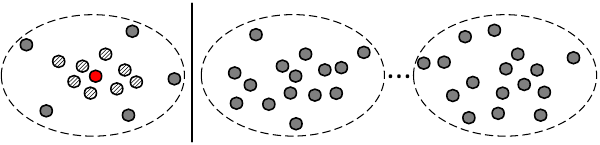}
	\caption{Red node is the anchor. Its negative sample set consists of dark-colored nodes from both within and outside the cluster. Shaded nodes are selected by the clustering refinement strategy and the self-correction mechanism, and they are likely to share the same true label as the anchor.}
	\label{tab:clustering-correction}
\end{figure}

Distant nodes are likely to serve as negative samples that enhance inter-class feature differences. Therefore, the distant nodes within the cluster are still selected as negative samples for the anchor node. In summary, for the text node $v_i$, its negative sample set(as illustrated in Fig \ref{tab:clustering-correction}) can be formalized as follows:
\begin{multline}
	N(v_i) = \{v_j \mid v_j \notin C(v_i)\} \cup {} \\
	\{v_j \mid v_j \in C(v_i) \land v_j \in distribution(v_i,d)\}
\end{multline}

\subsection{Model Training}
After optimizing the negative sampling through the clustering refinement strategy and the self-correction mechanism, the contrastive loss for any node \(v_i\) within view $1$ can be formalized as follows:
\begin{equation}
	\mathcal{L}_{v_i}^{1} = \frac{e^{Sim(z_i^{1},z_i^{2})/\tau}}{\sum_{v_j \in N(v_i)}e^{Sim(z_i^{1},z_j^{k})/\tau}}
\end{equation}
where $\tau$ is the temperature parameter, $z_i^1$ denotes the node representation of node $v_i$ in view $1$, and $k \in \{1, 2\}$. The function $Sim(\cdot,\cdot)$ represents the cosine similarity metric employed to quantify the resemblance between a pair of vectors. The overall contrastive loss function can be expressed as follows:
\begin{equation}
	\mathcal{L}_{cl} = \frac{1}{|V_1| + |V_2|} \sum_{i=1}^{|V_1|}(\mathcal{L}_{v_i}^{1} + \mathcal{L}_{v_i}^{2})
\end{equation}

Additionally, we utilize labelled samples to compute the cross-entropy loss for training the model. The text representations $Z_0$ that are learned on the original text graph through formula A are fed into the softmax classifier for label prediction. Formally,
\begin{equation}
	P = \textbf{softmax}(Z_0)
\end{equation}
The computation of the cross-entropy loss can be formalized as follows:
\begin{equation}
	\mathcal{L}_{ce} = - \sum_{i=1}^{|Y_L|} \sum_{j=1}^{M} Y_{ij}lnP_{ij}
\end{equation}

The final loss function $\mathcal{L}$, which directs the model's backpropagation, is mathematically formulated by integrating the contrastive loss with the cross-entropy loss as follows:
\begin{equation}
	\mathcal{L} = \mathcal{L}_{ce} + \beta\mathcal{L}_{cl}
\end{equation}

\section{EXPERIMENTS}
\label{sec:experrments}

\subsection{Datasets}
To evaluate the effectiveness of our proposed ClusterText in text classification tasks, we conducted extensive experiments on five commonly used datasets: 20Newsgroups, R8, R52, Ohsumed and Movie Review. Table \ref{tab:datasets} summarizes the statistical data for these datasets.

\begin{table}
	\centering
	\caption{THE STATISTICS OF FIVE DATASETS.}
	\label{tab:datasets}
	\resizebox{\columnwidth}{!}{%
		\begin{tabular}{lcccccc}
			\noalign{\hrule height 1pt}
			\noalign{\vskip 1pt}
			Dataset & \# Docs & \# Train & \# Label.rate & \# Classes & \# Avg.length \\
			\hline
			20NG & 18,846 & 11,314 & 60\% & 20 & 221.26 \\
			R8 & 7,674 & 5,485 & 71\% & 8 & 65.72 \\
			R52 & 9,100 & 6,532 & 71\% & 52 & 69.82 \\
			Ohsumed & 7,400 & 3,357 & 15\% & 23 & 135.82 \\
			MR & 10,662 & 7,108 & 66\% & 2 & 20.39 \\
			\noalign{\hrule height 1pt}
		\end{tabular}%
	}
\end{table}

\subsection{Baselines}
We consider five different types of text classification methods as baselines, including two GNN-based methods (TextGCN\cite{yao2019graph}, DADGNN\cite{liu2021deep}), a pre-trained method (BERT), and two GCL-based methods (CGA2TC\cite{yang2022contrastive}, TextGCL\cite{yang2022contrastive}).

\subsection{Experiment Setting}
For the clustering layer, we employ K-means clustering with the number of clusters set to the actual number of classes. It is important to note that the reported experimental results are the average of 10 runs with different weight initializations.

\subsection{Semi-supervised text classification performance}
\begin{table}[h]
	\centering
	\caption{TEST ACCURACY RESULTS (\%) FOR SEMI-SUPERVISED TEXT CLASSIFICATION OF DIFFERENT MODELS ON FIVE DATASETS.}
	\label{tab:model_comparison}
	\resizebox{\columnwidth}{!}{%
		\begin{tabular}{lccccc}
			\noalign{\hrule height 1pt}
			\noalign{\vskip 1pt}
			Model & 20NG & R8 & R52 & Ohsumed & MR \\
			\hline
			TextGCN & 86.3 & 97.1 & 93.6 & 68.4 & 76.7 \\
			DADGNN & 88.6 & 98.1 & 95.2 & 69.7 & 78.7 \\
			BERT & 85.3 & 97.8 & 96.4 & 70.5 & 85.7 \\
			CGA2TC & - & 97.8 & 94.8 & 71.0 & 78.0 \\
			TextGCL & 90.2 & 98.2 & 96.8 & 73.3 & 86.2 \\
			ClusterText & \textbf{90.7} & \textbf{98.6} & \textbf{97.4} & \textbf{74.0} & \textbf{87.5} \\
			\noalign{\hrule height 1pt}
		\end{tabular}%
	}
\end{table}
As illustrated in Table \ref{tab:model_comparison}, there is a notable disparity in performance among different models in the semi-supervised text classification task. Overall, our method consistently outperforms the baselines across all datasets. While traditional GNN-based models like TextGCN and DADGNN exhibit some efficacy in text classification, their performance is not optimal. These approaches rely solely on labelled samples to compute cross-entropy loss and do not leverage the rich, unlabeled data (test set) to enhance the model's training process. BERT achieves commendable classification results but does not consistently outperform the GNN-based methods on certain datasets. This may be due to BERT's insufficient utilization of inter-textual relational information. In contrast, CGA2TC and TextGCL, by integrating graph neural networks and contrastive learning, make effective use of both labelled and unlabeled samples, yielding better performance across multiple datasets compared to TextGCN, DADGNN, and BERT. However, their results still fall short of our proposed ClusterText. Our method achieves accuracy improvements of 0.5, 0.4, 0.6, 0.7, and 1.3 across different datasets. These results demonstrate that clustering pseudo-labels enhances the accuracy and robustness of semi-supervised text classification tasks.

\begin{table}[h]
	\centering
	\caption{TEST ACCURACY RESULTS (\%) UNDER DIFFERENT CLUSTERING METHODS.}
	\label{tab:varing-clustering-method}
	\resizebox{\columnwidth}{!}{%
		\begin{tabular}{lccccc}
			\noalign{\hrule height 1pt}
			\noalign{\vskip 1pt}
			Model & 20NG & R8 & R52 & Ohsumed & MR \\
			\hline
			K-means & 89.3 & 97.8 & 97.1 & 73.4 & \textbf{87.5} \\
			Agglomerative & \textbf{90.7} & \textbf{98.6} & 96.2 & 73.0 & 87.2 \\
			BRICH & 89.6 & 97.2 & \textbf{97.4} & \textbf{74.0} & 86.8 \\
			\noalign{\hrule height 1pt}
		\end{tabular}%
	}
\end{table}

\subsection{Results under Different Clustering Methods}
The optimization of negative sampling is based on clustering algorithms. Therefore, we evaluated ClusterText using K-means, AgglomerativeClustering, and BIRCH. It is worth noting that the initial number of clusters defaults to the true number of categories in the dataset. The experimental results are shown in Table \ref{tab:varing-clustering-method}. When using AgglomerativeClustering, ClusterText exhibited the best accuracy on the 20NG and R8 datasets. Conversely, on the R52 and Ohsumed datasets, ClusterText achieved the best results with the BIRCH algorithm. The variation range of results with different clustering algorithms is between 0.7 and 1.4, indicating that ClusterText has good adaptability to different clustering algorithms.

\subsection{Ablation Studies}
\begin{table}[h]
	\centering
	\caption{THE ABLATION RESULTS (\%) OF VARIOUS EXPERIMENTAL SETTINGS.}
	\label{tab:ablation_studies}
	\resizebox{\columnwidth}{!}{%
		\begin{tabular}{lccccc}
			\noalign{\hrule height 1pt}
			\noalign{\vskip 1pt}
			Model & 20NG & R8 & R52 & Ohsumed & MR \\
			\hline
			ClusterText & \textbf{90.7} & \textbf{98.6} & \textbf{97.4} & \textbf{74.0} & \textbf{87.5} \\
			w/o correction & 90.4 & 98.4 & 96.9 & 73.5 & 87.2 \\
			w/o clustering & 89.6 & 97.7 & 96.1 & 72.9 & 87.0 \\
			w/o GCL & 89.2 & 97.3 & 95.8 & 72.0 & 86.2 \\	
			\noalign{\hrule height 1pt}
		\end{tabular}%
	}
\end{table}

To validate the effectiveness of the proposed ClusterText, we conducted a series of ablation experiments on the 20NG, R8, R52, Ohsumed, and MR datasets. As shown in Table \ref{tab:ablation_studies}, the complete ClusterText method achieved the best performance. When modifications were made to ClusterText, accuracy declined across all datasets. Specifically, when the self-correction mechanism was removed, classification accuracy decreased by between 0.2 and 0.5, indicating that ignoring intra-cluster nodes in negative sampling reduces the number of true negative samples. On the other hand, removing the clustering refinement strategy resulted in a classification accuracy decrease ranging from 0.5 to 1.1. When contrastive learning was not considered, text classification performance also declined.

\section{CONCLUSION}
\label{sec:conclusion}
This paper proposes ClusterText, a cluster-optimized graph contrastive learning method for text classification. Specifically, ClusterText leverages both Bert and GCN to optimize text representation learning. Next, we introduce a clustering refinement strategy to mitigate negative sampling bias. Moreover, we propose a self-correction mechanism to identify true negative nodes within the same cluster as the anchor node, addressing the negative impact of clustering bias. Extensive empirical evaluations validate the superior efficacy of our proposed method in addressing semi-supervised text classification challenges. In future research, we will further explore GCL and scalable computing technologies, delving into their potential for large-scale text classification.

\bibliographystyle{IEEEtran}
\bibliography{refer}

\end{document}